# Impact of Dataset on Acoustic Models for Automatic Speech Recognition

Siddhesh Singh

Siddhesh Singh is in Department of Computer Science and Engineering, Manipal University Jaipur, India (e-mail: siddheshsingh123@gmail.com).

**ABSTRACT**

In Automatic Speech Recognition, GMM-HMM had been widely used for acoustic modelling. With the current advancement of deep learning, the Gaussian Mixture Model (GMM) from acoustic models has been replaced with Deep Neural Network, namely DNN-HMM Acoustic Models. The GMM models are widely used to create the alignments of the training data for the hybrid deep neural network model, thus making it an important task to create accurate alignments. Many factors such as training dataset size, training data augmentation, model hyperparameters, etc., affect the model learning. Traditionally in machine learning, larger datasets tend to have better performance, while smaller datasets tend to trigger over-fitting. The collection of speech data and their accurate transcriptions is a significant challenge that varies over different languages, and in most cases, it might be limited to big organizations. Moreover, in the case of available large datasets, training a model using such data requires additional time and computing resources, which may not be available. While the data about the accuracy of state-of-the-art ASR models on open-source datasets are published, the study about the impact of the size of a dataset on acoustic models is not readily available. This work aims to investigate the impact of dataset size variations on the performance of various GMM-HMM Acoustic Models and their respective computational costs.

## 1. INTRODUCTION

Automatic Speech Recognition has always been considered as an important junction in fostering better human-to-human and human-to-machine communication. From simple voice search on mobile to possibly making a speech-to-speech model for removing the language barrier, Automatic Speech Recognition is increasingly integrated into our daily lives. With the advancement of computational power today, it has become possible to train even more complex and robust models, opening paths for new possibilities.

In the past years, there has been research and development in many core technologies: Mel-Frequency Cepstral Coefficients (MFCCs) which are still widely used as features in ASR, Hidden Markov Models (HMM), Gaussian Mixture Models (GMM), Language Models such as n-grams. GMM-HMM Acoustic Models were widely used in ASR. They have contributed significantly to the advancement of state-of-the-art performance in ASR. With the advancement of computation, deep learning techniques have far surpassed the traditional state-of-the-art performance significantly- more than 1/3rd error rate reduction [12]. In the hybrid DNN-HMM implementations of Automatic Speech Recognition, GMMs are used to create alignments of the training data for these deep neural networks. With the increase in data and computational power, end-to-end deep neural networks are starting to show promising results.

In traditional deep learning networks, it is widely believed that using a larger train dataset would improve accuracy. However, obtaining large audio data with their correct transcriptions is not easy and might require many resources such as manual recording, making transcriptions of the available recordings, etc. In particular languages, large audio datasets might not be available. For the available large audio datasets, training models require huge computational resources. To train models with limited data or limited resources, it is essential to understand the tradeoff between the data size, computational cost, and model performance.

In end-to-end deep learning models [3, 11], there is a significant improvement with an increase in the dataset size. Deep Speech has noted that the Word Error Rate (WER) decreases by 40% (relative) upon the increase in train set size by a factor of 10 [1]. We want to analyze the impact of dataset scaling on the computational requirements and the WER on GMM-HMM Acoustic Models.

## 2. METHODOLOGY

Amongst several automatic speech recognition toolkits like

Kaldi [6], RWTH [7], HTK [8], Sphinx [10], etc., we have chosen Kaldi- an open-source Speech Recognition Toolkit which is written in C++ for training our models in the experiments. It consists of various useful tools that are used collectively to build a speech recognition system.

To analyze the impact of dataset scaling, computational requirement, and the WER on the GMM-HMM Acoustic Model, we randomly sampled three datasets of different sizes from the primary dataset. We perform a total of 5 Experiments of varying dataset sizes. In each experiment, 4 Models are trained iteratively on different partitioned datasets.

In the following subsections, we will discuss the computational specs, dataset selection and partitioning protocol, the acoustic models, and the training procedure

**2.1 Computational Specs**

All the models are trained on AWS EC2 instance c5a.24xlarge having 96 Core CPUs with 196 GiB Memory (RAM) on Ubuntu Server 18.04 LTS.

**2.2 Dataset**

Open-source English audio dataset LibriSpeech [5] contains 960 Hrs of speech data from audiobooks with a sample rate of 16 kHz. The training data is divided into three parts: train-clean-100, train-clean-360, and train-other-500, where total clean training data is 460 hrs and other data is 500 hrs. The testing data is divided into test-clean and test-other. For our experiment, the whole training set is merged, resulting in 960 hrs of training data, and the testing set is merged to form 10.5 hrs of testing data.

In this experiment, 960 Hrs training data is sampled randomly to create 4 Datasets- 50 Hrs, 100 Hrs, 500 Hrs, and 960 Hrs. Each of the sampled datasets is further sampled into three parts: Part 1 containing ~5% of data containing only the shortest utterances, Part 2 containing ~10% of the data, and Part 3 containing ~50% of the data. The following Table 1 depicts the size of the sampled datasets.

**Table 1.** Sampled Dataset Size.

| Sampled Dataset Size (Hrs) | Part 1 (5%) (Hrs) | Part 2 (10%) (Hrs) | Part 3 (50%) (Hrs) |
|---|---|---|---|
| 50 | 2.5 | 5 | 25 |
| 100 | 5 | 10 | 50 |
| 500 | 25 | 50 | 250 |
| 960 | 48 | 96 | 480 |

.

**2.3 Training**

Two types of Acoustic Models are trained- Monophones and Triphones. Monophone models compare each of the phonemes and assign them weights based on the match probability. Due to the comparison of each phone individually, Monophones do not capture context information. Triphone models are used to capture the context information- left and right phones. The captured phonetic context in Triphone models helps to tackle the effect of co-articulation [4] which Monophones do not handle.

The triphones have three types of training models: 1) Triphone 1 (Delta + Delta - Delta) [9], 2) Triphone 2 (LDA - MLTT), and 3) Triphone 3 (LDA + MLTT + SAT) [2]. These models are trained iteratively in the following manner-

a) The Monophone model is trained and is used to create alignments for Triphone 1.

b) The Triphone 1 model is trained using the alignments created by the Monophone model. It is used to create alignments for Triphone 2.

c) The Triphone 2 model is trained using the alignments created by the Triphone 1 model. It is used to create alignments for Triphone 3.

d) Finally, the Triphone 3 model is trained using the alignments created by the Triphone 2 model.

A total of 5 experiments are conducted. Each experiment contains models trained on different sizes particular to the sampled subset. In Experiment 0, all the models of the iteration, namely- Monophone, Triphone 1, Triphone 2, and Triphone 3 are trained on a complete dataset of 960 Hrs. In Experiment 1 - 4, Monophone is trained on Part 1 (~5% data) of the subset only containing shortest utterances; Triphone 1 is trained on Part 2 (~10% of data) of subset, Triphone 2 is trained on Part 3 (~50% data) of subset and Triphone 3 is trained on 100% of the data in its subset. The following Table 2 depicts the dataset size used in each experiment.

**Table 2.** Training Data Size for each Experiment.

| Exp. No. | Monophone (Hrs) | Triphone 1 (Hrs) | Triphone 2 (Hrs) | Triphone 3 (Hrs) |
|---|---|---|---|---|
| 0 | 960 | 960 | 960 | 960 |
| 1 | 2.5 | 5 | 25 | 50 |
| 2 | 5 | 10 | 50 | 100 |
| 3 | 25 | 50 | 250 | 500 |
| 4 | 48 | 96 | 480 | 960 |

This is done in order to understand two effects: a) Impact of using complete training data and only a subset of it in the initial training models of the iteration (like Monophone, Triphone 1, and Triphone 2). b) Impact of the dataset size on the final WER.

**3. RESULTS**

Table 3 shows the Word Error Rates (WER) of the experiments on each model trained iteratively, along with the time taken to train each model. In the following subsections, the experimental results are presented for the Acoustic Models on different sizes of the training data. It is to be noted that all the training has been done parallelly on 96 core CPUs with 96 parallel jobs.

**3.1 Impact of using a subset of training data on initial acoustic models**

In Experiment 0, the full training data (960 Hrs) is used for all the models of the training iteration- Monophone, Triphone 1, Triphone 2, and Triphone 3. In Experiment 1-4, only a subset of the full training data is used in the initial models of the iteration (Monophone, Triphone 1, and Triphone 2). 960 hours of training data is used to train the Triphone 3 model in both- Experiment 0 and Experiment 4, as shown in Table 2.

Table 3. WER and Training Time Taken for each Experiment.

| Exp. No. | Model Type | Training Data Size (Hrs) | WER % | Training Time (Minutes) | Total Training Time (Minutes) |
|---|---|---|---|---|---|
| 0 | Monophone | 960 Hrs | 56.98 | 62 | 149 |
|   | Triphone 1 | 960 Hrs | 31.61 | 19 |   |
|   | Triphone 2 | 960 Hrs | 25.4 | 30 |   |
|   | Triphone 3 | 960 Hrs | 20.66 | 38 |   |
| 1 | Monophone | 2.5 Hrs | 59.02 | 0.3 | 12 |
|   | Triphone 1 | 5 Hrs | 39.68 | 1.3 |   |
|   | Triphone 2 | 25 Hrs | 29.88 | 4.5 |   |
|   | Triphone 3 | 50 Hrs | 22.16 | 6 |   |
| 2 | Monophone | 5 Hrs | 57.99 | 0.5 | 15 |
|   | Triphone 1 | 10 Hrs | 35.5 | 1.5 |   |
|   | Triphone 2 | 50 Hrs | 27.47 | 5 |   |
|   | Triphone 3 | 100 Hrs | 21.29 | 8 |   |
| 3 | Monophone | 25 Hrs | 57.49 | 1.3 | 38 |
|   | Triphone 1 | 50 Hrs | 32.32 | 2.3 |   |
|   | Triphone 2 | 250 Hrs | 25.7 | 11 |   |
|   | Triphone 3 | 500 Hrs | 20.61 | 23 |   |
| 4 | Monophone | 50 Hrs | 57.4 | 3 | 61 |
|   | Triphone 1 | 120 Hrs | 31.76 | 4 |   |
|   | Triphone 2 | 480 Hrs | 25.51 | 17 |   |
|   | Triphone 3 | 960 Hrs | 20.64 | 38 |   |

From Table 3, it is observed that the WER of all the initial models of the training iteration of Experiment 4 (Monophone- 57.4%; Triphone 1- 31.76%; Triphone 2- 25.51%) is greater than that of Experiment 0 (Monophone- 56.98%; Triphone 1- 31.61%; Triphone 2- 25.4%). The major factor contributing to these differences is the training data size. However, the final WER is slightly lower in the case of Experiment 4 (Triphone 3- 20.64%) than that of Experiment 0 (Triphone 3- 20.66%), even though training data for Triphone 3 of both the experiments are the same (960 Hrs). Even though the word error rates of both the experiments are very similar, the total training time taken by them vastly differs due to a decrease in the training data size of the initial models of the training iteration. In Experiment 0, the total training time is 149 minutes (~2.48 Hrs), while the total training time of Experiment 4 is 61 minutes (~1 Hr). Due to subsetting the dataset for initial models, along with a decrease in WER, the training time is decreased by more than half of the initial training time.

### 3.2 Impact of the dataset size

In Experiments 1-4, a subset of training data as shown in Table 2 is used to train the initial models of the training iteration. Total training data for Experiment 1 is 50 Hrs, Experiment 2 is 100

Hrs, Experiment 3 is 500 Hrs and Experiment 4 is 960 Hrs. In Experiment 1-3, the final WER decreases with an increase in the training data size.

In Experiment 4 (960 Hrs), the final WER slightly increases as compared to Experiment 3 even though the WER of the initial models of Experiment 3 has lower WER than that of experiment 4.

With the ~7% decrease (relative) of WER in Experiment 4 where 960 hrs of training data is used when compared to Experiment 1 where 50 hrs of training data is used, the training time (and computational cost) increases by more than 4 times. Such similar observations from Table 3 can be used to make decisions on the tradeoff of word error rates and computational cost for a particular use case of the Automatic Speech Recognition System.

## 4. CONCLUSION

In recent years, the usage of Automatic Speech Recognition System services has increased. However, the success of such services is dependent on the availability of datasets and computational power. Collecting speech data may be challenging for particular languages, and resources for training massive datasets might not be available. This work aims to investigate the impact of dataset size on the performance of GMM-HMM Acoustic Models. For this purpose, we carried out five experiments on different sizes of a dataset.

Two observations can be made from the experiments performed. a) Training the models with the full training data might not necessarily be beneficial. Our experiment has shown that using only a subset of the full training data to train the initial models (Monophone, Triphone 1, and Triphone 2) not only gives an improvement in WER but reduces the training time significantly. b) The final WER falls with an increase in the dataset size, and the results of the experiments can be used to decide the right size of the dataset to train the Acoustic Model. These results are highly useful in determining the tradeoff of data size, training time, and accuracy of the model even before training. In case of limited computational resources, this tradeoff can be considered for the training depending on the WER requirements and availability of computational resources.

A natural progression of this research would be to investigate the performance of other available datasets of different sizes on Acoustic Models- GMM-HMM and DNN-HMM and study the impact of the total number of trainable parameters of these models on different training datasets.

## ACKNOWLEDGMENT

This work was sponsored by ClusterDev Technologies, Kerala 682021, India.

## REFERENCES

1. Amodei, D., Ananthanarayanan, S., Anubhai, R., Bai, J., Battenberg, E., Case, C., Casper, o.: Deep speech 2: End-to-end speech recognition in english and mandarin. In: International Conference on Machine Learning (2016).
2. Guglani, J. and Mishra, A.N. DNN based continuous speech recognition system of Punjabi language on Kaldi toolkit. International Journal of Speech Technology, 24, pp.41-45 (2021).
3. Hannun, Awni, Case, Carl, Casper, Jared, Catanzaro, Bryan, Diamos, Greg, Elsen, Erich, Prenger, Ryan, Satheesh, Sanjeev, Sengupta, Shubho, Coates, Adam, et al.: Deepspeech: Scaling up end-to-end speech recognition. arXiv preprint arXiv:1412.5567 (2014).
4. Knill, K. and Young, S., 1997. Hidden Markov models in speech and language processing. In Corpus-based methods in language and speech processing. Springer, Dordrecht (1997) pp. 27-68.
5. Panayotov, V., Chen, G., Povey, D., Khudanpur, S.: Librispeech: an asr corpus based on public domain audio books. In: IEEE international conference on acoustics, speech and signal processing (ICASSP). IEEE, pp 5206–5210 (2015).
6. Povey, D., Ghoshal, A., et al.: The Kaldi speech recognition toolkit. In: ASRU. IEEE Signal Processing Society (2011).
7. Rybach, D., Gollan, C., Heigold, G., Hoffmeister, B., Lööf, J., Schlüter, R., Ney, H.: The RWTH aachen university open source speech recognition system. In: Interspeech, Brighton, U.K. 2111–2114 (September 2009).
8. S. Young, G. Evermann, M. Gales, T. Hain, D. Kershaw, X. Liu, G. Moore, J. Odell, D. Ollason, D. Povey, V. Valtchev, and P. Woodland. The HTK Book. Cambridge University Engineering Department, HTK version 3.4 edition, December (2006).
9. Tokuda, K., Kobayashi, T., & Imai, S. Speech parameter generation from hmm using dynamic features. In: ICASSP, pp. 660–663 (1995).
10. Walker, W., Lamere, P., Kwok, P., Raj, B., Singh, R., Gouvea, E., Wolf, P. and Woelfel, J.: Sphinx-4: a flexible open source framework for speech recognition. Technical Report SMLI TR2004-0811, Sun Microsystems, Inc. (2004).
11. Watanabe, S., Hori, T., Karita, S., Hayashi, T., Nishitoba, J., Unno, Y., Soplin, NEY., Heymann, J., Wiesner, M., Chen, N., et al: ESPnet: end-to-end speech processing toolkit. In: Proceedings of Interspeech, Hyderabad (2018).
12. Yu, D., & Deng, L.: Automatic speech recognition (Signals and Communication Technology). Springer, London (2015).